\documentclass[10pt,twocolumn,letterpaper]{article}

\usepackage{cvpr}
\usepackage{times}
\usepackage{epsfig}
\usepackage{graphicx}
\usepackage{subcaption}
\usepackage{amsmath}
\usepackage{amssymb}

\usepackage[breaklinks=true,bookmarks=false]{hyperref}

\cvprfinalcopy 

\def\cvprPaperID{****} 

\begin{document}

\title{EmotioNet Challenge: Recognition of facial expressions of emotion in the wild}

\author{C. Fabian Benitez-Quiroz, Ramprakash Srinivasan, Qianli Feng, Yan Wang, Aleix M. Martinez\\
Dept. Electrical and Computer Engineering\\
The Ohio State University
}

\maketitle

\begin{abstract}
This paper details the methodology and results of the EmotioNet challenge. This challenge is the first to test the ability of computer vision algorithms in the automatic analysis of a large number of images of facial expressions of emotion in the wild. The challenge was divided into two tracks. The first track tested the ability of current computer vision algorithms in the automatic detection of action units (AUs). Specifically, we tested the detection of 11 AUs. The second track tested the algorithms' ability to recognize emotion categories in images of facial expressions. Specifically, we tested the recognition of 16 basic and compound emotion categories. The results of the challenge suggest that current computer vision and machine learning algorithms are unable to reliably solve these two tasks. The limitations of current algorithms are more apparent when trying to recognize emotion. We also show that current algorithms are not affected by mild resolution changes,  small occluders, gender or age, but that 3D pose is a major limiting factor on performance. We provide an in-depth discussion of the points that need special attention moving forward. \end{abstract}

\section{Introduction}

Much progress has been made in computer vision in the last couple decades. Of late, tremendous efforts have been made to design algorithms that can detect and recognize generic and specific object categories in images, with tremendous achievements recorded just in the last few years. ImageNet, PASCAL-VOC and COCO are three example databases and challenges \cite{Russakovsky:14,Everingham:10,Lin:14} that have helped fuel the progress and success of these efforts.

In the present paper, we ask if these achievements extend to the detection and recognition of facial expressions of emotion {\em in the wild}. Previous studies have reported good results in the automatic analysis of facial expressions of emotion \cite{Corneanu:16}. However, these results were obtained with the analysis of images and videos taken in the laboratory \cite{Zafeiriou:16,Dhall:16}. That is, even when the expressions were spontaneous, the filming was done in controlled conditions with the full awareness of the participants. Also, researchers had access to both the training and testing data, allowing practitioners to modify their algorithms until these work on the known testing data.

To evaluate whether current computer vision algorithms can fully automatically analyze images of facial expressions of emotion in the wild, we collected a large database of 1 million image \cite{Benitez-Quiroz:16}. This datase is called EmotioNet. We also developed a computational model based on the latest knowledge of how the human brain analyzes facial expressions \cite{Martinez:17}. This algorithm was used to annotate a subset of 950,000 images of the EmotioNet set. The remaining 50,000 images were manually labelled by expert annotators. 

The 950k images that were automatically annotated by our algorithm defined the training set. The training set was made available to participants of the challenge. 25k of the manually annotated images were also made available to participants as a validation set. This was necessary because the images in the training set are not accurately annotated, i.e., the labels are unreliable. The accuracy of annotations in the training set is about 81\%. The validation set can thus be used to determine how well one's algorithm adapts to this unreliability. 

The final 25k of the manually annotated images are part of the testing set. This is a sequestered set, i.e. not available to teams participating in the challenge during training and algorithm development. The testing set was the only one used to provide the final evaluation of participating teams.

The results of the challenge show that current computer vision algorithms cannot reliably detect and recognize facial expressions of emotion. In fact, our analysis of the results suggest much progress is needed before computer vision algorithms can be deployed in uncontrolled, realistic conditions. For example, 3D pose is shown to be a major factor in performance -- in general, the more a face deviates from frontal view, the less accurate the results are. 

Nevertheless, there are positive results to report too. For example, small local occluders and image resolution (scale) were not found to affect the accuracy and robustness of the tested algorithms. This is a significant achievement, since just a decade back, occlusions and scale were major factors in face recognition tasks \cite{Zhao:16,Liu:15,Jia:09,Kotsia:08}. Also, we demonstrate that neither gender nor age have an effect on detection of AUs or recognition of emotion categories.

\section{Definition of the challenge}

Facial expressions of emotion are produced by contracting one's facial muscles \cite{Martinez:17}. Muscle groups that lead to a clearly visible image change are called action units (AUs). Each of these AUs is associated with a unique number. For example, AU 1 specifies the contraction of the frontalis muscle pars medialis, and AU 2 defines the contraction of the same muscle pars lateralis. These result in the upper movement of the inner and outer corners of the eyebrows, respectively \cite{Ekman:06}. 

The first task in the EmotioNet challenge was to identify eleven different AUs. These AUs and a description of their visible actions are given in Table \ref{Table: AUs}. 

\begin{table}
\begin{center}
\begin{small}
\begin{tabular}{|l|l|l|l|}
\hline
AU \# & Action & AU \# & Action\\
\hline
1 & inner brow raiser & 2 & outer brow raiser\\
\hline
4 & brow lowerer & 5 & upper lid raiser\\
\hline
6 & cheek raiser & 9 & nose wrinkler\\
\hline
12 & lip corner puller & 17 & chin raiser\\
\hline
20 & lip stretcher & 25 & lips part\\
\hline
26 & jaw drop & -- & --\\
\hline
\end{tabular}
\end{small}
\end{center}
\caption{ The 11 AUs used in track 1 of the challenge. Listed here are the 11 AUs alongside the definitions of their actions.}\label{Table: AUs}
\end{table}

\begin{figure}
\centering
\includegraphics[width=.15\textwidth]{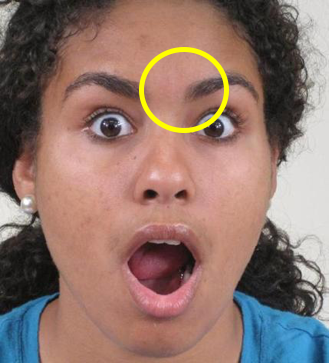}
\includegraphics[width=.15\textwidth]{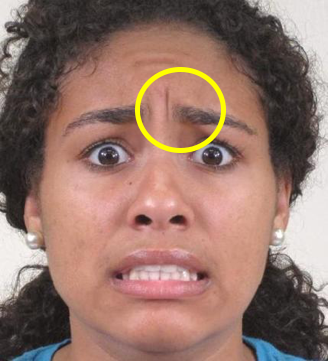}
\includegraphics[width=.15\textwidth]{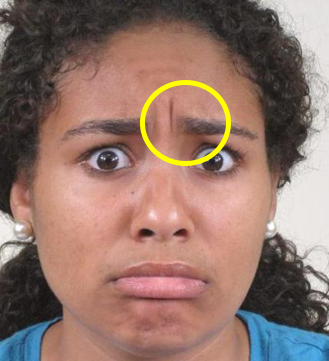}
\includegraphics[width=.15\textwidth]{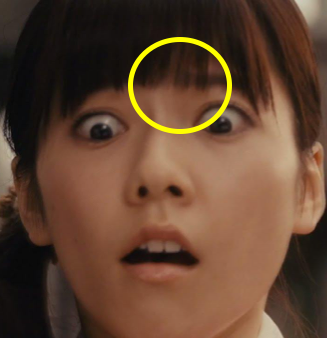}
\includegraphics[width=.15\textwidth]{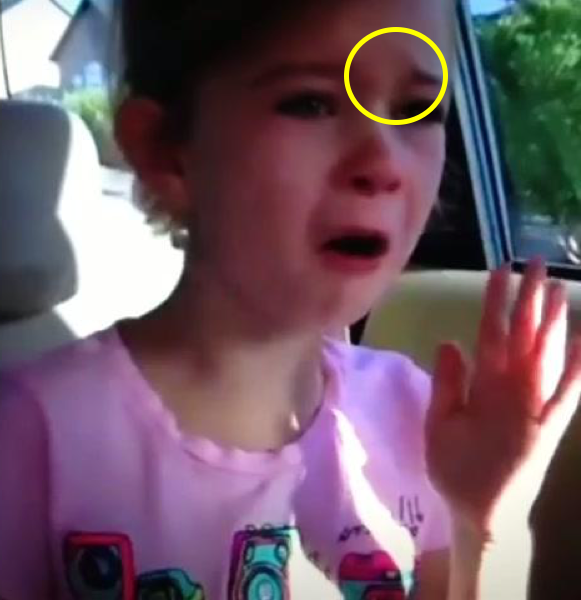}
\includegraphics[width=.15\textwidth]{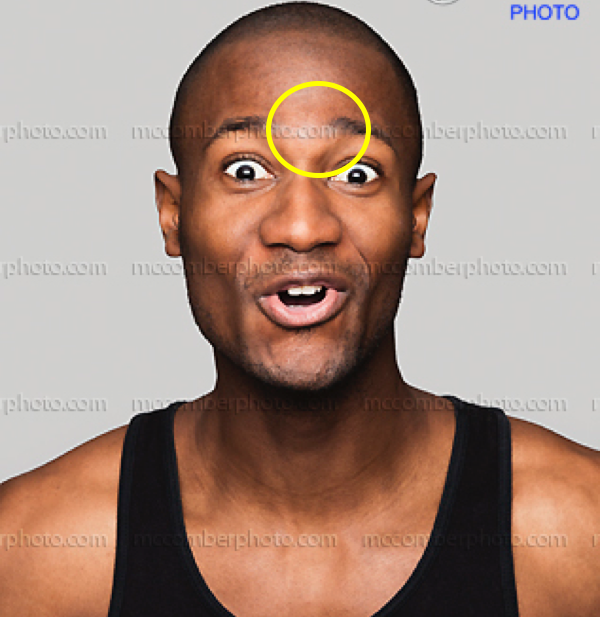}
\caption{This figure illustrates the heterogeneity of shape and shading features associated to the same AU. The images above all have AU 1 active, yet the image features that define this activation are quite different from image to image. Yellow circles indicate the area effected by the activation of AU 1.}\label{Fig: AU 1}
\end{figure}

This is a challenging task, because the appearance of an AU varies significantly when used in conjunction with other AUs. Figure \ref{Fig: AU 1} illustrates this. The top three images in this figure show facial expressions performed by the same person. These images were taken in a controlled environment with a cooperative subject. The three expressions have AU 1 present; shown within a yellow circle. Note how different the local shape and shading within the yellow circles are. This problem is only exacerbated in images of facial expressions of emotion in the wild where the heterogeneity of identity, ethnicity, illumination and pose is much larger. This is shown in the bottom three images in Figure \ref{Fig: AU 1}. Computer vision and machine learning algorithms need to learn to detect AU 1 in all these instances. 

Successful computer vision algorithms need to learn to identify image features that are exclusively caused by the activation of an AU. This task is so difficult that expert human coders have to go through intensive training and practice sessions. The ones that successfully complete their training are called expert coders. That was the problem addressed in track 1 of the challenge: Can we design a computer vision system that is as good as human expert coders?

The second track of the challenge went a step further by requiring computer vision algorithms to recognize sixteen possible emotion categories in facial expressions. The 16 emotion categories are: happy, angry, sad, surprised, fearful, disgusted, appalled, awed, angrily disgusted, angrily surprised, fearfully angry, fearfully surprised, happily disgusted, happily surprised, sadly angry, and sadly disgusted. The first seven of these categories are typically referred to as basic emotions. The others are called compound emotions \cite{Du:14}.

Emotion categories are easily defined as having a unique set of AUs. That is, the AUs must be the same (consistent) for expressions of the same emotion category {\em and} differential between emotion categories. The AUs defining the sixteen emotions included in track 2 of this challenge are given in Table \ref{Table: Emotion Categories}.

\begin{table}\scriptsize
\begin{center}
\begin{small}\footnotesize
\begin{tabular}{|l|l|l|l|}
\hline
Category & AUs & Category & AUs\\
\hline
Happy &  12, 25  & Sadly disgusted &   4, 10   \\
\hline
Sad &  4, 15     & Fearfully angry &  4, 20, 25   \\     
\hline
Fearful &  1, 4, 20, 25    & Fearfully surpd. &  1, 2, 5, 20, 25 \\
\hline
Angry &  4, 7, 24  & Sadly angry  &  4, 7, 15 \\
\hline
Surprised& 1, 2, 25, 26 & Angrily surprised &   4, 25, 26 \\
\hline
Disgusted &  9, 10, 17   & Appalled &  4, 9, 10 \\
\hline
Happily surpd. &  1, 2, 12, 25 & Angrily disgusted &  4,  10, 17 \\
\hline
Happily disgd. &   10, 12, 25   & Awed &   1, 2, 5, 25 \\
\hline
\end{tabular}
\end{small}
\end{center}
\caption{Prototypical AUs used to produce each of the sixteen basic and compound emotion category of track 2 of the challenge.}\label{Table: Emotion Categories}
\end{table}

\section{Methodology}

With regard to detection of AUs and recognition of emotion categories, there are several evaluation criteria that need to be considered. 

The first one is accuracy. Accuracy measures the number of true and false positives. This is important because we wish to know if our algorithm is able to discriminate between sample images with a certain AU/emotion present. In statistics, this is generally called observational error -- the difference between the measured and true value. In other words, we wish to know the bias of an algorithm.

A second important criterion is precision. Precision, which is also known as positive predictive value, is the fraction of correctly identified instances of an AU or emotion category. Statistically, this measures the extent to which the distribution of errors is stretched.

A third and final criterion to consider is recall. Recall, also called sensitivity, is the number of correctly detected/recognized instances of a class over the true number of samples in that class. In other words, a highly sensitive test, rarely overlooks a true positive.  

Formally, accuracy is defined as
\begin{equation}
\text{accuracy}_i = \frac{\text{true positives}_i + \text{true negatives}_i}{\text{total population}},
\end{equation}
where $i$ specifies the class, i.e., AU $i$ or the $i^{th}$ emotion category, $\text{true positives}_i$ are correctly identified test instances of class $i$, $\text{true negatives}_i$ are test images correctly labeled as not belonging to class $i$, and $\text{total population}$ is the total number of test images.

Precision is given by $\text{precision}_i = \text{detected}_i/\text{true}_i$, where $\text{detected}_i$ is the number of images where our algorithm has detected (or recognized) class $i$, and $\text{true}_i$ is the actual number of images belonging to that class.

And, recall is $\text{recall}_i = \text{correct}_i/\text{true}_i$, with $\text{correct}_i$  the number of correctly detected/recognized images of class $i$.

A convenient, highly employed criterion that efficiently combines precision and recall is the F$_\beta$ score, $\beta>0$. The F$_\beta$ of class $i$ is defined as,
\begin{equation}
\text{F}_{\beta_i} = \left(  1+\beta^2\right) \frac{\text{precision}_i \text{recall}_i}{\beta^2 \text{precision}_i + \text{recall}_i}.
\end{equation}

Distinct values of $\beta$ are used to measure different biases and limitations of an algorithm. For example, $\beta=.5$ gives more importance to precision. This is useful in applications where false positives are not as important as precision. In contrast, $\beta=2$  emphasizes recall, which is important in applications where false negatives are unwarranted. And, $\beta=1$  provides a measure where recall and precision are equally relevant. 

The present challenge used the values of accuracy and F$_1$ score to evaluate participating algorithms. The final score of the challenge was computed as the average of these two criteria,
\begin{equation}\label{Eq: Final score}
\text{final score} = \frac{\text{accuracy} + \text{F}_1}{2},
\end{equation}
where $\text{accuracy}=C^{-1}\sum_{i=1}^C \text{accuracy}_i$, $\text{F}_1= C^{-1}\sum_{i=1}^C \text{F}_{1_i}$, and $C$ is the number of classes. We also provide the results of F$_\beta$, with $\beta=.5$ and $2$. But the winner of the challenge was given by the final score defined in (\ref{Eq: Final score}).

\section{Training and testing sets}

The challenge was divided into two tracks. {\bf Track 1} required participants to detect the presence of 11 AUs in images of facial expressions. These AUs were listed in Table \ref{Table: AUs}. Training data was made available to participants in November of 2016. The training dataset consists of 950k AU-annotated images of facial expressions. The AU annotations are given by the algorithm of \cite{Benitez-Quiroz:16}. The accuracy of these annotations is about 81\%.

Participants of track 1 were given access to a verification set. The verification set consists of 25k images of facial expressions with {\em manual} AU annotations. These annotations were given by expert human coders. Annotations were cross referenced by fellow coders for verification of accuracy. 

\begin{figure}
\centering
\includegraphics[width=.15\textwidth]{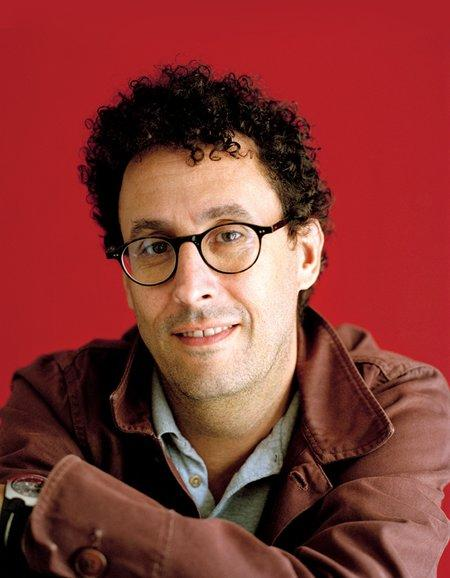}
\includegraphics[width=.15\textwidth]{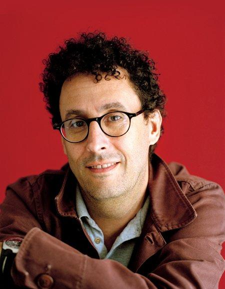}
\includegraphics[width=.15\textwidth]{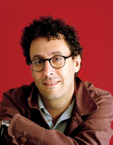}
\caption{Shown here are three test images in (left to right) their original scale (resolution) as well as at 1/2 and 1/4 of the original scale.}\label{Fig: Scales}
\end{figure}

{\bf Track 2} of the challenge required participants to recognize one of 16 possible emotion categories. The emotion categories included in this challenge are listed in Table \ref{Table: Emotion Categories}. Participants had access to the training data of track 1 which includes many sample images with the emotion categories included in this track. Table \ref{Table: Emotion Categories} specifies which AU combinations correspond to each of the emotions of the challenge.

Additionally, participants of this track were given a validation set with verified expressions of each of the sixteen emotions in Table \ref{Table: Emotion Categories}. The validation set includes 2k sample images. This validation set was made available to participants in December of 2016.

{\bf Testing set (track 1).} The testing sets were sequestered and were not available to participants during training and algorithm design.

The testing data of track 1 included 88k images. Of those, 22k were images with manual annotations given by expert human coders. An additional 44k were obtained by reducing the resolution of these 22k images. This was done by reducing the image to 1/2 and 1/4 of their original size. This yielded images with decreasing resolution (scale). Figure \ref{Fig: Scales} shows an example image at these three scales.

\begin{figure}
\centering
\includegraphics[width=.155\textwidth]{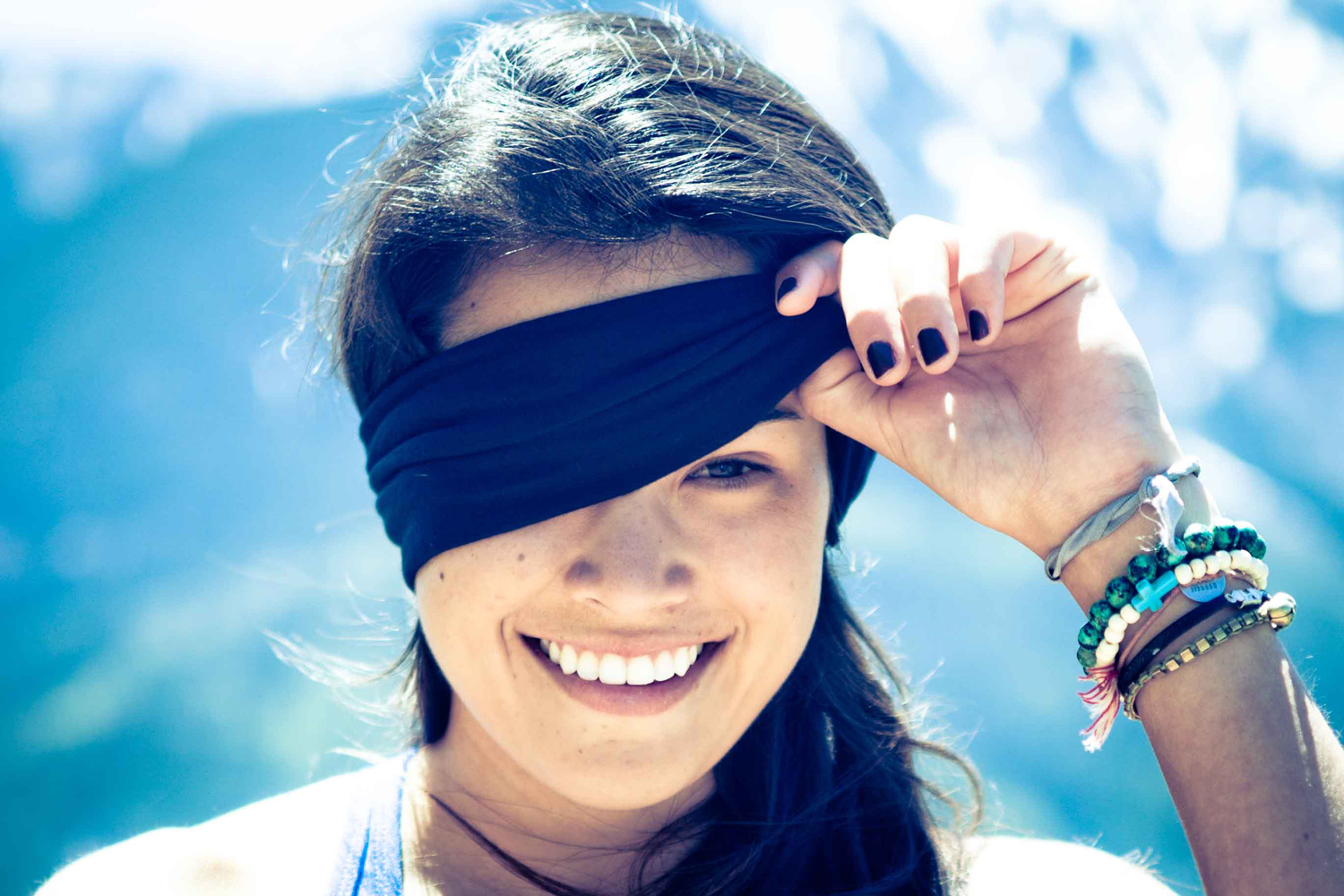}
\includegraphics[width=.165\textwidth]{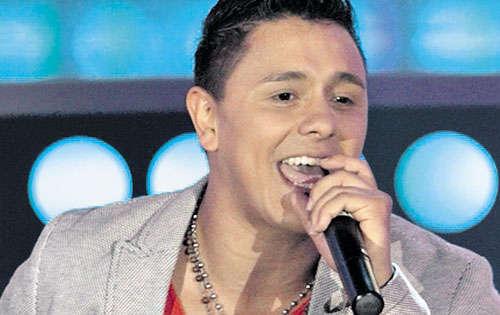}
\includegraphics[width=.133\textwidth]{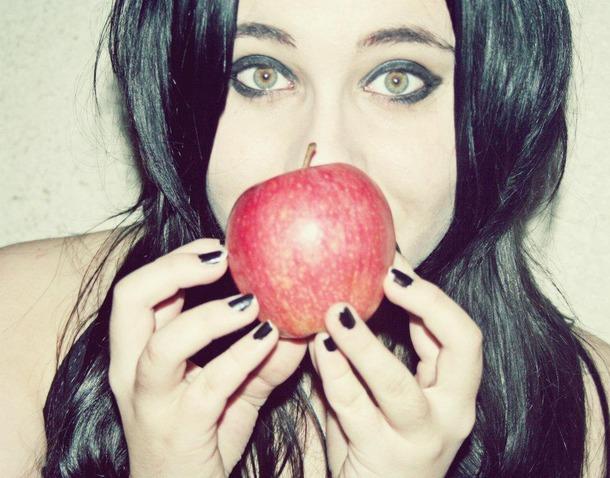}
\caption{Three example images with natural occlusions.}\label{Fig: Natural occlusions}
\end{figure}

Another 22k were obtained by adding small occluding black squares in the images. Occlusions were added in areas of the face that did {\em not} occlude the local region of AU activation. This was done to determine whether current algorithms are unaffected by small distractors. Occlusions were black squares of size equal to 1/5$^{th}$ the width of the face. Face width was defined as the distance between the far-most corners of the left and right ears. Note we only tested for small unnatural distractors. This test is not meant to test for large artificial occluders. Nonetheless, natural occluders are common in our training and testing sets. Figure \ref{Fig: Natural occlusions} shows a few examples of natural occlusions found in EmotioNet.

{\bf Testing set (track 2).} The testing set of track 2 included 40k images. Of these, 10k were images of facial expressions of the sixteen emotions listed in Table \ref{Table: Emotion Categories}. An additional 20k were obtained by reducing the size of these images by 1/2 and 1/4, as above. And a final 10k corresponded to the same images with the addition of the small occluders described in the preceding paragraph.

{\bf Challenges of track 1 versus track 2.} The main challenges in track 1 are: to be able to deal with the heterogeneity of the images in EmotioNet, and to detect the small (sometimes barely visible) image changes caused by some of the tested AUs. On the latter point, note that while some AUs lead to large image changes, others yield hardly visible ones. For instance, AU 12 results in clearly visible mouth changes, taking a large area of the face. In contrast, AU 26 can yield tiny image changes that are hard to learn to detect.

The challenges of track 2 are compounded by those of track 1. Recall that an emotion category is defined by having a set of AUs. This was shown in Table \ref{Table: Emotion Categories}. Thus, unless one can solve the problem of AU detection, it would be very difficult to recognize emotion. 

One solution to the above problem is to define an algorithm that learns to recognize emotion categories, rather than AUs. The problem is that the results of such an algorithm correlate with those of AU detection, as demonstrated by \cite{Du:14}. Also, it is believed that the human brain first interprets AUs, before making a determination of the emotion displayed in the face \cite{Martinez:17}. To this point, a neural mechanisms involved in the decoding of AUs in the human brain has been identified \cite{Srinivasan:16}. These results suggests that the best approach to recognize emotion is to first detect AUs.

Thus, overall, the problem defined in track 2 is considered significantly more challenging than that of track 1.

\section{Results}

To participate in this challenge, online registration was mandatory. Detailed instructions on how to participate were only sent to groups that had completed the registration process.

Thirty-eight (38) groups registered to participate. These groups received final instructions on how to evaluate their algorithms on the sequestered testing sets in early February of 2017. Participants had a few days to complete their tasks. 

Of the original 38 groups, only 5 submissions were successfully completed before the assigned deadline. We received a couple of requests after the deadline, but these were not included in our evaluation given below.

\subsection{Track 1}

Table \ref{Table: Results track 1} shows the results of the final scores of the four submissions we received in track 1 before the deadline. The algorithm at the top of this table obtained the highest final score, as given by (\ref{Eq: Final score}). This was the winner of track 1.  

\begin{table}
\begin{center}
\begin{small}
\begin{tabular}{|l|l|l|l|}
\hline
Algorithm & final score $\uparrow$  & accuracy & F$_1$ score\\
\hline
I2R-CCNU-NTU-2 & {\bf .729} & .822 & .64\\
\hline
JHU & .71 & .771 & .64\\
\hline
I2R-CCNU-NTU-1 & .702 & .784 & .63\\
\hline
I2R-CCNU-NTU-3 & .696 & .776 & .622\\
\hline
\end{tabular}
\end{small}
\end{center}
\caption{Main results of track 1. Note that final score takes a value between 0 and 1, with 0 being the worst and 1 the best.}\label{Table: Results track 1}
\end{table}

Results of baseline algorithms are provided in Table \ref{Table: Baseline results track 1}. These include results obtained using: {\em a.} Kernel Subclass Discriminant Analysis (KSDA) \cite{You:11}, a standard probabilistic method that, and {\em b.} AlexNet \cite{Krizhevsky:12}, a standard deep learning convolutional neural network. In both cases, we trained a classifier for each of the AUs. This yielded 11 two-class classifiers, each able to detect one of the 11 tested AUs. Only the training set was used to compute these classifiers.

The four participating algorithms are convolutional neural networks. The top algorithm uses residual blocks and a sum of binary cross-entropy loss. The second algorithm from the top uses a multi-label softmax loss function \cite{Wang:17}.

\begin{table}
\begin{center}
\begin{small}
\begin{tabular}{|l|l|l|l|}
\hline
Algorithm & final score $\uparrow$  & accuracy & F$_1$ score\\
\hline
KSDA & .708 & .807 & .615\\
\hline
AlexNet & .608 & .828 & .388\\
\hline
\end{tabular}
\end{small}
\end{center}
\caption{Baseline algorithm results for track 1, AU detection.}\label{Table: Baseline results track 1}
\end{table}

Table \ref{Table: F scores track 1} shows the F$_{.5}$ and F$_2$ scores of the four participating algorithms plus the two baseline methods introduced in the preceding paragraph.

\begin{table}
\begin{center}
\begin{small}
\begin{tabular}{|l|l|l|}
\hline
Algorithm & F$.5$ score $\uparrow$  & F$_2$ score\\
\hline
I2R-CCNU-NTU-2 & .64 & .643 \\
\hline
JHU & .638 &  .635\\
\hline
I2R-CCNU-NTU-1 & .635 &  .625\\
\hline
I2R-CCNU-NTU-3 & .627 &  .62\\
\hline
\hline
KSDA & .621 &  .611\\
\hline
AlexNet & .353 & .446 \\
\hline
\end{tabular}
\end{small}
\end{center}
\caption{F$_{.5}$ and F$_2$ scores of the algorithms tested in track 1.}\label{Table: F scores track 1}
\end{table}

\begin{table}
\begin{center}
\begin{small}
\begin{tabular}{|l|l|l|l|}
\hline
Algorithm & final score $\uparrow$  & accuracy & F$_1$ score\\
\hline
I2R-CCNU-NTU-2 & .729 & .822 & .641\\
\hline
JHU & .702 & .763 & .632 \\
\hline
I2R-CCNU-NTU-1 & .699 & .782 & .626\\
\hline
I2R-CCNU-NTU-3 & .694 & .775 & .62\\
\hline
\hline
KSDA & .71 & .807 & .619\\
\hline
AlexNet & .515 & .763 & .266\\
\hline
\end{tabular}
\end{small}
\end{center}
\caption{Results on the set of images with different scales and small unnatural occluders of track 1.}\label{Table: Scales occluders track 1}
\end{table}

And, Table \ref{Table: Scales occluders track 1} shows the final scores, accuracies and F$_1$ scores for the images with varying scale and small artificial occluders.

The evaluations shown in Tables \ref{Table: Results track 1}-\ref{Table: Scales occluders track 1} indicate that most of the tested algorithms yield similar results. Crucially, these results have much room for improvement. The final scores for example are all below $.73$, and the F$_\beta$ scores are $<.65$, significantly below the maximum attainable value, which is 1.

\subsection{Track 2}

Table \ref{Table: Results track 2} provides the final scores, accuracies and F1 scores of the two submissions we received before the deadline plus a baseline algorithm for this track. 

\begin{table}
\begin{center}
\begin{small}
\begin{tabular}{|l|l|l|l|}
\hline
Algorithm & final score  & accuracy & F$_1$ score\\
\hline
NTechLab & {\bf .597} & .941 & .255\\
\hline
JHU & .48 & .836 &  .142\\
\hline
\hline
KSDA & .578 & .91 & .247\\
\hline
\end{tabular}
\end{small}
\end{center}
\caption{Main results of track 2. The algorithm at the top of the table was the winner of this track.}\label{Table: Results track 2}
\end{table}
 
\begin{table}
\begin{center}
\begin{small}
\begin{tabular}{|l|l|l|}
\hline
Algorithm & F$.5$ score & F$_2$ score\\
\hline
NTechLab & .258 & .26\\
\hline
JHU & .182 & .127\\
\hline
\hline
KSDA & .274 & .242\\
\hline
\end{tabular}
\end{small}
\end{center}
\caption{F$_{.5}$ and F$_2$ scores of the algorithms tested in track 2.}\label{Table: F scores track 2}
\end{table}

\begin{table}
\begin{center}
\begin{small}
\begin{tabular}{|l|l|l|l|}
\hline
Algorithm & final score  & accuracy & F$_1$ score\\
\hline
NTechLab & .602 & .94 & .267 \\
\hline
JHU & .465 & .82 & .13 \\
\hline
\hline
KSDA & .532 & .88 & .2\\
\hline
\end{tabular}
\end{small}
\end{center}
\caption{Results on the set of images with different scales and small unnatural occluders of track 2.}\label{Table: Scales occluders track 2}
\end{table}

The baseline algorithm is defined as follows. First the KSDA baseline algorithm of track 1 is used to estimate the AUs present in the test images. Since KSDA is a probabilitic approach, the presence of each AU is given by a probability of detection. Let $p_i$ be the probability of detecting AU $i$. Then, the probability of detecting an emotion category is given by,
\begin{equation}
\text{emotion}_j = \sum_{\forall \text{AUs}} w_{ji}\, p_i.
\end{equation}
Here, $w_{ji}$ are the weights associated to each possible AU. These weights are learned using KSDA on the verification data. The final decision of this baseline algorithm is simply given by,
\begin{equation}
\arg\max_j \text{emotion}_j.
\end{equation}

Table \ref{Table: F scores track 2} shows the F$_{.5}$ and F$_2$ scores for the same three algorithms. And Table \ref{Table: Scales occluders track 2} has the results for the images of different scales (1/2 an 1/4 of the original size) and with small occluders.

Two conclusions must be drawn from these results. First, the F$_\beta$ scores for all algorithms are very low. This assessment illustrates the limitations of current computer vision and machine learning algorithms on the recognition of emotion categories from images. This problem is considered solved for images in controlled lab conditions \cite{Du:14,Corneanu:16}. But, the results of the EmotioNet Challenge demonstrate there is still much to be accomplished before these algorithms can be utilized in complex applications broadly.

Second, the results of this challenge shows that small artificial occluders and image resolution are {\em not} limiting factors in the classification of emotion. This points needs to be stressed, because only a few years ago this was still an open problem \cite{Zhao:16,Liu:15,Jia:09,Kotsia:08}.

\begin{figure*}
\centering
\begin{subfigure}[b]{0.24\textwidth}
	\includegraphics[width=\textwidth]{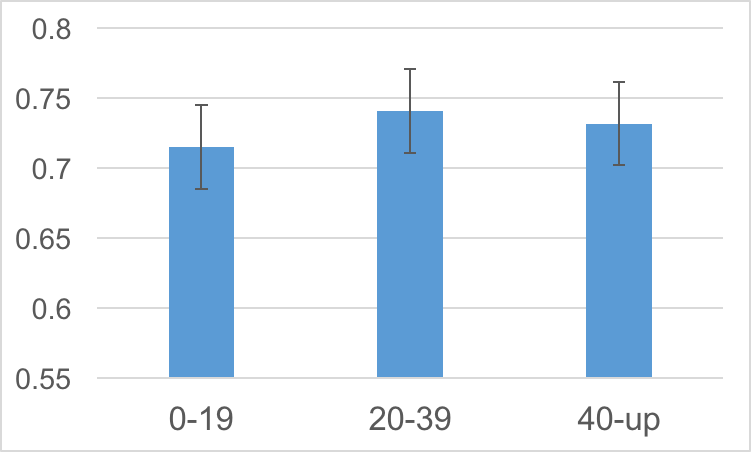}
	\caption{}\label{Figure a}
\end{subfigure}
\begin{subfigure}[b]{.24\textwidth}
	\includegraphics[width=\textwidth]{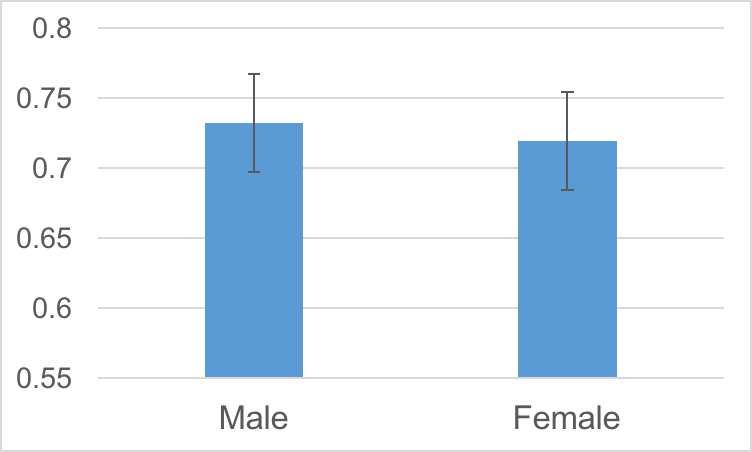}
	\caption{}\label{Figure a}
\end{subfigure}
\begin{subfigure}[b]{.24\textwidth}
	\includegraphics[width=\textwidth]{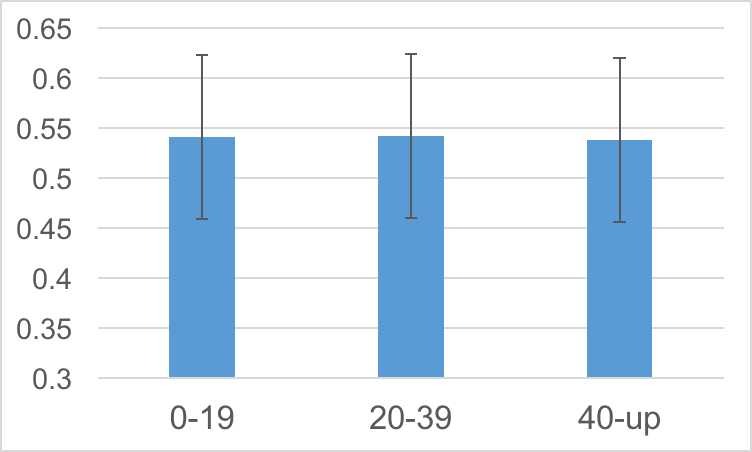}
	\caption{}\label{Figure a}
\end{subfigure}
\begin{subfigure}[b]{0.24\textwidth}
	\includegraphics[width=\textwidth]{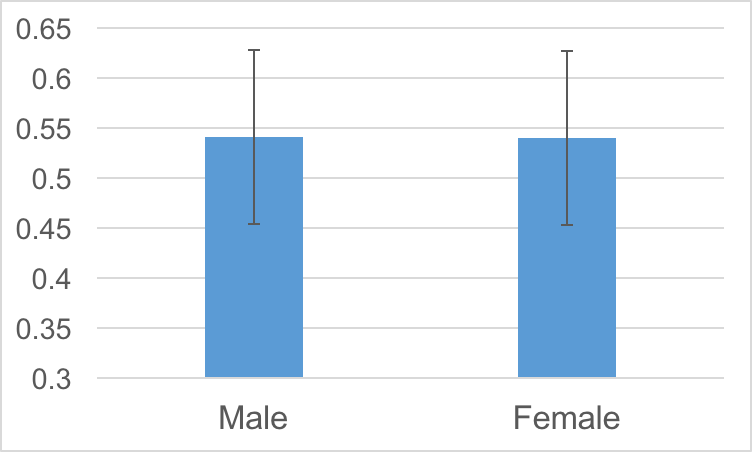}
	\caption{}\label{Figure a}
\end{subfigure}
\caption{Average and standard deviations of the final scores of participating algorithms in task 1 (AU detection) for: (a) the three age groups, and (b) the two gender groups. Average and standard deviations of the final scores of participating algorithms in task 2 (emotion recognition) for: (c) the three age groups, and (d) the two gender groups.}\label{Fig: Biases}
\end{figure*}

\subsection{No age or gender biases}

We also tested for possible biases in the database and algorithms participating in this challenge. 

First, we used the regression algorithm of \cite{You:14} to estimate the age and gender of the faces in the EmotioNet database. This regression method is a proven algorithm, yielding state-of-the-art results.

Age is divided into three groups: 0-19 years of age, 20-39, and 40 and above. The percentages of images in each of these three groups are: 37\%, 45\% and 18\%, respectively. Not surprisingly, images found online mostly represent young people.

Conversely, the percentage of images in each gender are: 48\% male and 52\% female. Thus, the database does not have any biases on gender.

Are any of the results described above biased by these percentages?

To test this, we rerun the detection and recognition results described above for each of these five groups (3 age groups and 2 gender groups). 

The results show no major biases due to either age or gender. For instance, in detection of AUs, the accuracy results of the I2R-CCNU-NTU-2 algorithm are:  .779 (in the 0-19 age group), .837 (in the 20-39 group) and .843 (in the 40-above group). The accuracy results of the JHU algorithm are: .734, .801, .764, respectively. The results of the I2R-CCNU-NTU-1 algorithm are: .773, .788, .786, respectively. And, the results of the I2R-CCRU-	NTU-3 algorithm are: .773, .788, .786, respectively. The F$_\beta$ scores are equally similar across age groups. Hence, these results only show a very mild disadvantage for the youngest faces (0-19 group).

In emotion recognition, the NTechLab accuracy results are: .938 (in the 0-19 group), .943 (in the 20-39 group) and .941 (in the 40 and above group). The JHU accuracy results are: .83 (in the 0-19 years group), .83 (in the 20-39 group) and .822 (in the 40-above group). The F$_\beta$ scores are equally similar to each other. Hence, no emotion classification biases were identified.

We found the same result in the gender groups, i.e, detection of AUs and recognition of emotion categories were unbiased by gender.

Figure \ref{Fig: Biases} summarizes the average results of all algorithms in each of these groups. We run four $t$-tests, one for each of the plots in this figure, and found no statistical difference between either age or gender ($p>.2$).

\subsection{Face pose is a major factor}

Finally, we assessed the algorithms' ability to deal with pose. That is, are the detection and recognition results reported above dependent on the pose of the face?

To study this question, we used the algorithm of \cite{Zhao:16} to compute the pose of all the faces of the EmotioNet database. This algorithm was a top performer in a recent competition of 3D pose estimation, the 2016 3D Face Alignment in the Wild Challenge \cite{Jeni:16}. Extensive evaluation shows that the error of this algorithm in computing the 3D pose of a face is  $<.004$ mm \cite{Ruiqi:16b}.

We use this algorithm to compute roll, pitch and yaw. Roll is given by the normal vector to the center point of the tip of the nose. Thus, only pitch are yaw define 3D distortions of the face. To see this, note that the in-plane rotation of roll does not affect the outcome of a computer vision algorithm, since any face detector will readily eliminate it. Hence, 3D effects are formally defined by,
\begin{equation}\label{Eq: pose}
\frac{|\text{pitch}|+|\text{yaw}|}{2},
\end{equation}
where $|.|$ is the absolute value.

\begin{figure}
\centering
\begin{subfigure}[b]{0.23\textwidth}
	\includegraphics[width=\textwidth]{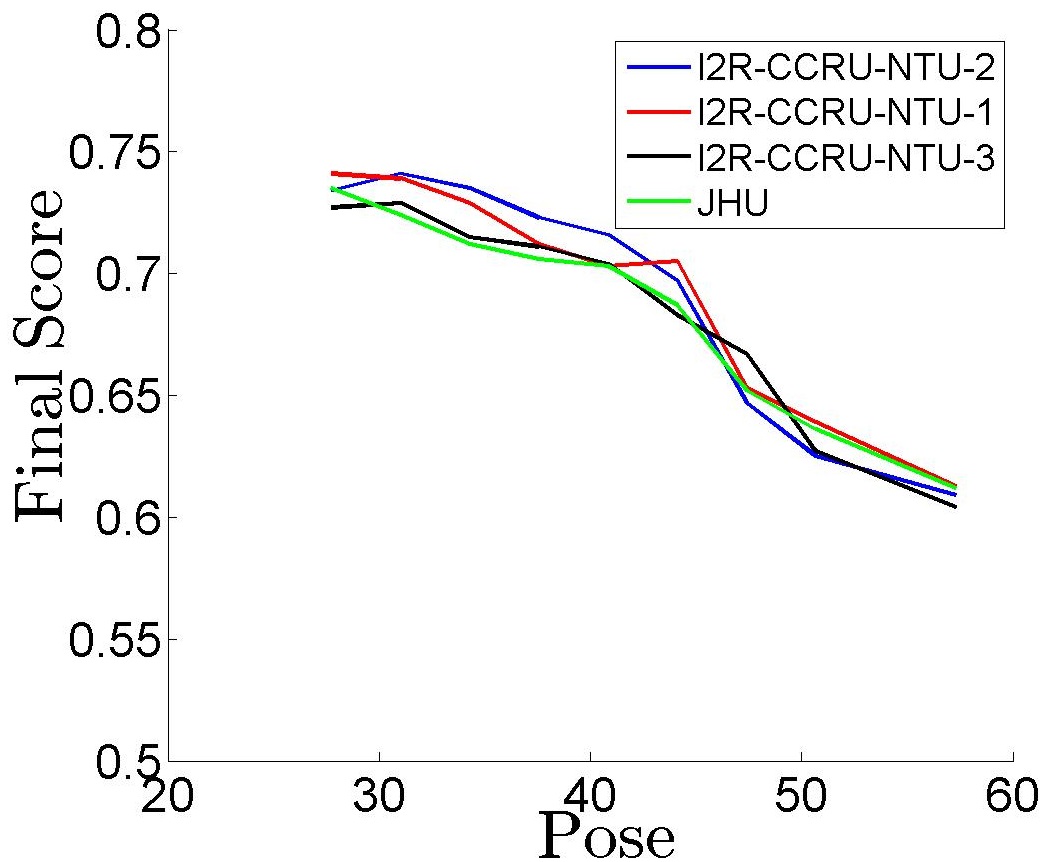}
	\caption{}\label{Figure a}
\end{subfigure}
\begin{subfigure}[b]{.23\textwidth}
	\includegraphics[width=\textwidth]{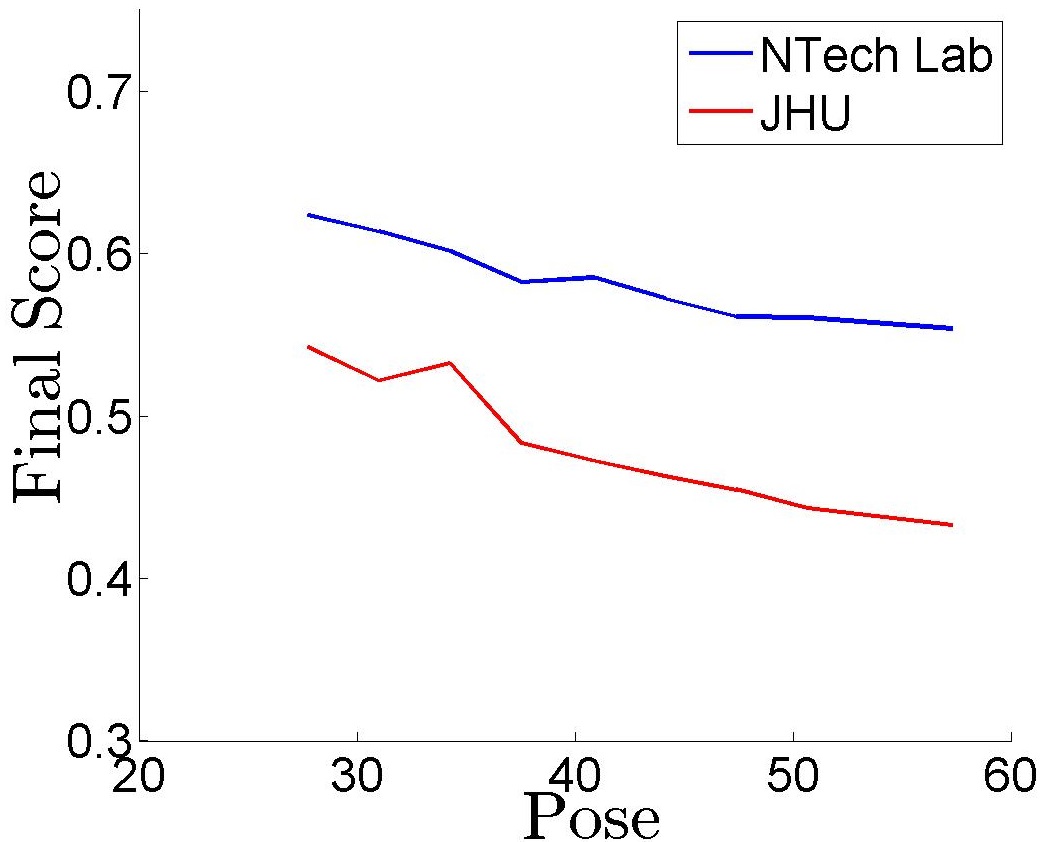}
	\caption{}\label{Figure a}
\end{subfigure}
\caption{Final score of participating algorithms as a function of pose: (a) AU detection, (b) emotion recognition. Pose is defined in equation (\ref{Eq: pose}) and is given in degrees in the $x$-axis. The $y$-axis is the final score given by equation (\ref{Eq: Final score}).}\label{Fig: Pose}
\end{figure}

Figure \ref{Fig: Pose} illustrates the effect that pose has on the final score of AU detection and emotion recognition for each of the participating algorithms.

As can be clearly seen in this figure, both, AU detection and emotion recognition, are significantly affected by pose. Clearly, this is an area that needs to be addressed if computer vision systems are ever to compete with human performance.

\section{Discussion}

This paper summarizes the results of the 2016-17 EmotioNet Challenge. This is the first ever challenge on the recognition of facial expressions of emotion in the wild. Key factors of the challenge are: the use of an unprecedentedly large number of images of facial expressions, the heterogeneity of expression in the wild, a large number of tested AUs {\em and} emotion categories, and the fact that testing was completed on a sequestered set of images not available to participants during training and algorithm development.

The most important result of the challenge is the realization of the difficulty of emotion categorization in expressions in the wild by current computer vision and machine learning algorithms. While the results make it clear that the detection of AUs need to be significantly improved, it is even more apparent that the recognition of emotion is not even close to being solved. Here, it is important to highlight the fact that the present challenge tested 16 emotions, more than doubling the number of categories used in previous experiments. Yet, additional emotion categories exist, as shown in \cite{Du:14,Benitez-Quiroz:16}. If current algorithms cannot accurately recognize 16 emotions, they will have an even harder time identifying additional ones.

EmotioNet provides a large number of labelled images. Hence, the limitations listed in the preceding paragraphs cannot be attributed to a lack of labelled data. Nonetheless, some of these labels were imprecise (noisy). Thus, a major open area of interest in computer vision and machine learning is how to learn from inaccurate labelled data. We believe that this can be accomplished with the help of the validation set, which is manually annotated and verified. Unfortunately, most current algorithms do not take advantage of this fact. 

One solution to the above problem is to learn co-activation patterns in the validation set and then impose this on the training set. Related methods were studied in the past (e.g., \cite{Tong:07}), but these approaches have been abandoned recently in favor of deep learning methods. Thankfully, these two approaches are not mutually exclusive and could be easily integrated.

A positive outcome of the challenge that deserve celebration is to note that scale and minor image distractors do not have a noticeable effect on detection and recognition. It is easy to dismiss this fact, but much effort has been made in the last fifteen years since these problems were introduced \cite{Martinez:02}. 

The present results show that the community is now ready to test detection and recognition in highly degraded images, e.g., tiny resolutions and major occlusions. The human visual system is robust to these image manipulations \cite{Martinez:17} and, hence, we know the problem is solvable.  

Also, an in-depth analysis of the results of participating algorithms showed no biases due to age or gender. This is also a positive outcome, not shared by all face recognition challenges \cite{Kemelmacher:16}.

However, all participating algorithms are influenced by the 3D pose of the face. Frontal faces are found to be much easier. And, in general, the accuracy of detection and recognition decrease as a function of pitch and yaw. This results calls for the need to normalize faces to a frontal view before applying classification algorithms. Indeed, this is an area of intense research \cite{Jeni:16,Ruiqi:16b,Tome:17}.

\section{Conclusions}

The EmotioNet Challenge was divided into two tracks. The first track tested computer vision algorithms' ability in detecting 11 AUs. This is a task successfully completed by human experts only. The second track assessed computer vision algorithms' ability to recognize 16 emotion categories. This is a task successfully completed by most people, experts and non-experts.

The images in the EmotioNet challenge are of facial expressions of emotion in the wild, Figures \ref{Fig: AU 1}-\ref{Fig: Natural occlusions}. This is a major departure from previous studies and challenges were images of facial expressions of emotion had been collected in controlled conditions in the lab. 

The results of the EmotioNet Challenge illustrate the difficulty of these two task in images of facial expressions of emotion in the wild. Nonetheless, a surprising and interesting outcome of the challenge is the extra difficulty algorithms have in successfully completing the task of the second track. 

Results of the first track, which can only be completed by expert human coders, show current algorithms are not fully ready. But, the recognition of emotion, which any layperson can successfully complete, is a much harder problem for computer vision algorithms. This is a well-known result in artificial intelligence (AI). AI systems tend to be better at solving human tasks that require expertise. And, the same algorithms are generally significantly worse at solving day-to-day tasks that humans take for granted.

In summary, the results of the studies delineated in the present paper, define a new frontier for computer vision systems. Research is needed to determine if current algorithms  need tweaking, or whether a novel set of algorithms is required to emulate these humans' abilites.


{\small
\bibliographystyle{ieee}
\bibliography{cvpr2016}
}

\end{document}